\documentclass[conference]{IEEEtran}
 \usepackage[pdftex]{graphicx}
\usepackage{algorithmic}

\usepackage{array}

\ifCLASSOPTIONcompsoc
  \usepackage[caption=false,font=normalsize,labelfont=sf,textfont=sf]{subfig}
\else
  \usepackage[caption=false,font=footnotesize]{subfig}
\fi
\usepackage{dblfloatfix}

\usepackage{url}

\begin{document}
%
\title{About Pyramid Structure in \\ Convolutional Neural Networks}



%

\author{\IEEEauthorblockN{Ihsan Ullah\IEEEauthorrefmark{1}\IEEEauthorrefmark{2} and
Alfredo Petrosino\IEEEauthorrefmark{2}}

\IEEEauthorblockA{\IEEEauthorrefmark{2}CVPR Lab, Department of Science and Technology, University of Napoli 'Parthenope',  Naples, Italy\\
Email: ihsan.ullah@unimi.it \/ alfredo.petrosino@uniparthenope.it}
\IEEEauthorblockA{\IEEEauthorrefmark{1}Department of Computer Science, University of Milan, 
Milan, Italy}}


\maketitle



\begin{abstract}
   Deep convolutional neural networks (CNN) brought revolution without any doubt to various challenging tasks, mainly in computer vision. However, their model designing still requires attention to reduce number of learnable parameters, with no meaningful reduction in performance. In this paper we investigate to what extend  \textit{CNN} may take advantage of pyramid structure typical of biological neurons. A generalized statement over convolutional layers from input till fully connected layer is introduced that helps further in understanding and designing a successful deep network. It reduces ambiguity, number of parameters, and their size on disk without degrading overall accuracy. Performance are shown on state-of-the-art models for \textit{MNIST, Cifar-10, Cifar-100}, and \textit{ImageNet}-12 datasets. Despite more than $80\%$ reduction in parameters for \textit{Caffe\_LENET}, challenging results are obtained. Further, despite $10-20\%$ reduction in training data along with $10-40\%$ reduction in parameters for \textit{Alexnet} model and its variations, competitive results are achieved when compared to similar well-engineered deeper architectures.
\end{abstract}


%
\IEEEpeerreviewmaketitle

\section{Introduction}\label{sec:Intro}
Convolutional Neural Networks (\textit{CNNs}) plays a significant role in raising deep learning society. However, all these models use the same concept of producing feature maps in convolutional layer followed by a pooling layer to reduce the dimension of the map.
Models such as \textit{Alexnet} \cite{Krizhevsky_imagenetclassification}, \textit{GoogleNet} \cite{SzegedyLJSRAEVR15}, \textit{VGG} \cite{Simonyan14c} and many others \cite{Jia:2014:CCA:2647868.2654889, icml2013_wan13, NIPS2014_5348, He2015} \cite{He_2015_ICCV} asserts that the deeper you go, the more the network performs well. In addition, going deeper the models slightly changed the concept for avoiding vanishing of gradients by using class inference in consecutive convolution layers and max pooling layer, or using a layer wise softmax layer that re-boosts the vanishing gradients \cite{SzegedyLJSRAEVR15, Simonyan14c,ZeilerF14}. Others used new activation functions, weight updates regularization methods, class inferences, layer wise pre-training in supervised approaches which showed very promising results \cite{icml2013_wan13,He_2015_ICCV}. Increasing number of layers means increasing number of parameters in a network. With the introduction of Network in Network (\textit{NIN}) model \cite{NINLinCY13}, the issue of reducing parameters is further highlighted once again. However, it tends toward greater computation.  As most of parameters exist in fully connected (\textit{FC}) layers, therefore, C. Szegedy et al. \cite{SzegedyLJSRAEVR15} use sparsity reduction complex methodologies for refining the trained models. S. Han et al. \cite{learningConnections15} tried to learn connections in each layer instead of weights and than retraining the network for reducing number of parameters.
Unfortunately, these aforementioned models are not suitable for real world mobile devices: since these  enourmously rises the computational operations, the number of parameters is unarguably a substantial issue in application space and increases the memory cost due to large size of trained models on the disk. \par
Biological studies lead to the idea of Image pyramids (IP's) \cite{nnvsimpyr}. IP's shown to be an efficient data and processing structure for digital images in a variety of vision applications e.g. object, digit recognition \cite{nnvsimpyr}. IP's as well as NN are massively parallel in structure. Pyramids at their simplest are like stack of filtered images with exponentially reduced dimensions.  
Pyramids have a long relation with machine learning and computer vision. Several models have been proposed based on the concept of pyramids e.g. Neocognitron, early LENET, Pyramidal neural network, spatial pyramids, and several others \cite{Lecun98gradient-basedlearning,Phung2007,Lazebnik2006}. In some recent works, pyramid structure is used only in one layer like the Spatial Pyramid Matching (SPM) \cite{Lazebnik2006} for pooling layer. H. Fan et al. \cite{DeepFaceFanCJYD14} tried to use pyramid structure in its last Conv layer of CNN for face recognition on LFW and showed 97.5\% accuracy. P. Wang et al. \cite{TempPyramidPoolWangCSLS15} utilizes pyramid structure better than the aforementioned work. They applied temporal pyramid pooling to enhance and use the temporal structure of videos just like spatial pyramids in \cite{StochasticPoolZeiler} where, they incorporated weak spatial information of local features. Their pyramidal temporal pooling method showed better results than the state-of-the-art two-stream model \cite{NIPS2014twostream} on HMDB1 dataset. On the other hand, models like \textit{PyraNet} \cite{Phung2007}, \textit{I-PyraNet} \cite{Fernandes2008}, \textit{LIPNET} \cite{Fernandes2013} and their extended models emphasized strict pyramidal structure from input till output. Their objective was to show that following strict pyramidal structure can enhance performance as compared to unrestricted models. \par
We have explored and proposed some basic hints about the main questions regarding CNN for image classification problem. 
The questions we have identified to answer are:
impact of reversing number of kernels in a convolutional layer?, does reversing a model work in every case?, impact of strict pyramidal structure?, and how to reduce number of parameters without complex rules and loss of accuracy?.  
To answer these questions we have utilized some well-known state-of-the-art models e.g. LENET \cite{Lecun98gradient-basedlearning}, AlexNet and its modified referenced model BVLC\_Reference\_Caffe \cite{Krizhevsky_imagenetclassification}. We have shown that the same number of filters can be used in different but pyramidal order without affecting the performance of the base network. \par
The rest of the paper is organized in four sections. In section \ref{sec:background}, a background of CNN and pyramidal models is provided. Section \ref{sec:proposedmodel} introduce strictly pyramidal structure of CNN's (SPyr\_CNN), while section \ref{resultexperiemts} tries to answer questions about architecture and performance with our experimental results that we have achieved for referenced \textit{CNN's} and \textit{SPyr\_CNN's}. Finally, in section \ref{concl}, a conclusion is presented with some future directions. 
\section{Background}\label{sec:background}
There are four key ideas behind CNNs that take advantage of the properties of an image: local connections, shared weights, pooling and use of many layers. Deep models or specifically CNNs exploit the compositional hierarchies of the images; in which higher-level features are obtained by composing lower-level ones. For example, local combinations of edges form motifs, motifs assemble into parts, and parts form objects.
The role of the convolutional layer is to detect local conjunctions of features from the previous layer i.e. edges, axons, etc.; whereas, pooling layers not only reduces dimensionality, but also merges semantically similar features into one. This helps in providing translation and scale invariance to small shifts and distortions. Y. Lecun's \cite{Lecun98gradient-basedlearning} early CNN followed strict pyramidal approach. In addition its weight sharing concept helped neural networks to reduce burden of large amount of trainable parameters. However, recent CNN's doesn't follow strict structure, although they reduced feature map size at each higher layer but increases number of maps as well which further increases total trainable parameters. Some of the well known models such as; AlexNet \cite{Krizhevsky_imagenetclassification} that won ILSVRC had 60M parameters, DeepFace model with 120M parameters resulted in best face recognition accuracy, or a recent very deep model \cite{Simonyan14c} that contains 133-144M parameters for getting about 24-30\% top-1 error on ILSVRC-12 dataset starts from $96$ feature maps and ends at 512 or more maps at a higher layer. \par
Therefore, some recent works like \cite{MemBoundCNNCollinsK14} and \cite{NINLinCY13} highlighted the issue
of memory usage in deep networks by reducing number of parameters. M. Lin et al. \cite{NINLinCY13} proposed a unlikelier local patch modeling in \textit{CNN} by replacing linear convolutions in each layer with convolving the input with a micro-network filter. This micro filter works like a multi-layer perceptron. This technique was extended and used in inception model \cite{SzegedyLJSRAEVR15} as a micronetworks modules. It works as dimensionality reduction to remove computational bottlenecks and reduce storage costs. 
Collins and Kohli \cite{MemBoundCNNCollinsK14} used sparsity-inducing regularizers during training to encourage
zero-weight connections in the \textit{Conv} and \textit{FC} layers. 
T. Sanath et al. \cite{conf/icassp/SainathKSAR13} exploited low-rank matrix factorization to reduce network parameters in FC layers. 
M. Denil et al. \cite{NIPS2013_5025} tried to predict parameters from other parameters.  \par
Strict pyramidal models \cite{Phung2007, FUKUSHIMA1988119, hubel1963} takes large amount of data as an input, refine it layer by layer in-order to reduce unwanted features and to enhance the final decision based on most likely and less number of unambiguous features. This gave base to many feature extraction and selection process in CV and ML. 
In computer vision, SPM with SIFT+FV  technique dominated ILSVRC classification and detection competitions until AlexNet arrived \cite{Krizhevsky_imagenetclassification}. 
Lazebnik et al. \cite{Lazebnik2006} introduces spatial pyramids technique which were mostly used for object recognition before deep CNN's. J. Masci et al. \cite{Masci2012} used multi-scale pyramid pooling layer to get a fixed size input feature vector. Recently, a much deeper model was introduced having inspiration from both \cite{Lazebnik2006} and \cite{Masci2012}. This spatial pyramid pooling \textit{(SPP)} \cite{He2015} approach provides multiple fixed size inputs to \textit{FC} layer with the help of pyramid pooling, which showed better results than normal fixed size input models e.g. Overfeat, AlexNet etc. 
\section{Pyramidal structure in CNNs}\label{sec:proposedmodel}


Despite success of \textit{CNN}, there is no principled way of finding good performing \textit{CNN} architectures other than naive exploration of the architecture space. This is a question that many newcomers ask: how to arrange filters in each layer for modeling a good network? Less reasoning has been given other than; increasing number of maps as the network go deeper will give good results. Therefore, we have introduced simple rule of thumb when designing a \textit{CNN} architecture based on the study of pyramidal neurons in the brain and human nature, highlighting this main question \par
\textit{''If a deep CNN architecture with specific number of layers and filters works on a specific task, while keeping the number of layers same and reversing the number of filters, provided that it forms pyramidal structure, the resultant pyramidal architecture will result in same or better performance?''} \par
\begin{figure}[t]
\centering
   \includegraphics[width=3.5in]{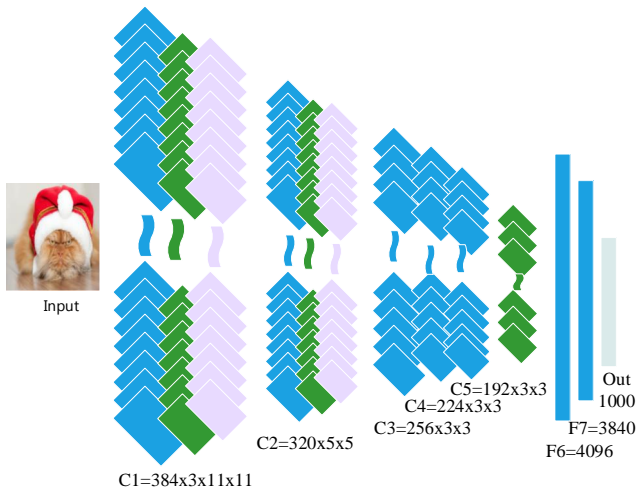}

   \caption{Strictly Pyramidal Architecture for CNN (SPyr\_CNN).}
\label{fig:S_Pyr_CNN}
\end{figure}

This question highlights two main aspects. We ask for if a network still works, if reversed, and,  secondly, if it works also with a pyramidal structure.
In addition to pyramidal structure, hints about this question is taken from a trick in PCA approach, where only changing the order of the matrices in calculating covariance matrix, not only reduce the time complexity but also avoided memory overflow issue \cite{PCA1991}. Answering this question thus helped in proposing a general rule of modeling a network by reducing a constant factor $'f'$ of filters from layer $'l'$ compare to previous layer $'l-1'$, other than better understanding the network behavior.  
Thus, we have investigated a strict pyramid network \textit{SPyr\_CNN} architecture as shown in Fig.~\ref{fig:S_Pyr_CNN}. It starts from a big input/first layer and then refine the features at each higher layer, until it achieves a reduced most discriminative set of features as being done in \cite{Phung2007, Fernandes2008, Fernandes2013}. Imposing strict pyramidal structure should not only retain and improve accuracy, but also retain/improve accuracy and reduce the number of parameters which results in less number of memory space on the disk. And this makes CNNs more feasible for applications where there is lack of memory. 
To achieve this goal, we will present experimental evidences  based on comparison with state-of-the-art models and datasets in the following section. 
\section{Experimental Results}\label{resultexperiemts} 
We will provide empirical evidence that a \textit{SPyr\_CNN} on small to large benchmark datasets \textit{MNIST, CIFAR-10, CIFAR-100} and \textit{ImageNet-12} performs similarly or better than complex state-of-the-art deep CNN models despite reduction in parameters.
\textit{MNIST} is a well-known 10 class hand written digits recognition dataset \cite{lecun-mnisthandwrittendigit-2010}. It consists of 60000 training and 10000 testing images. 
\textit{Cifar-10} is an object recognition dataset \cite{Krizhevsky09learningmultiple}. It consists of 60000 colour images with 6000 images per class. In each class 5000 images are used for training where as remaining 1000 images are used for testing. We have used these datasets to perform an in depth study of different models based on comparison with state-of-the-art, since the CNN can be easily trained and tested with reasonable computing and time cost. 
\textit{Cifar-100} is also an object recognition dataset having 100 classes but same number of images as Cifar-10 \cite{Krizhevsky09learningmultiple}. However, due to less images per class that results in less training data, it is considered  medium size challenging dataset.     
\textit{ImageNet}-2012 dataset consist of 1000-classes \cite{ILSVRC15}. It comprises of 1.2 million training images and 50,000 validation images (used as testing set). The results are measured by top- 1 and top-5 error rates. We have only used the provided data for training; all the results were evaluated on the validation set. Top-5 error rate is the metric officially used to rank the methods in the classification challenge.  
We perform two different experiments in order to evaluate power of pyramidal networks, i.e. with full training sets and with reduced training sets. Hence we reduced 10\% and 20\% randomly selected images from Cifar-100 and ILSVRC-12 training dataset. 
To generalize a common rule of selecting number of filters in each layer of a deep network, we used Cifar-10 and ILSVRC-12 for examining the best decrement factor for a successful model. 
These models will be explored in the up-coming sections.
Strict pyramidal models will be represented by \textit{SPyr} in beginning of each model. We have implemented these \textit{SPyr\_CNN} models on widely deployed Caffe \cite{Jia:2014:CCA:2647868.2654889} framework. The hyper-parameters for training setup are all same as Alexnet model and its variant BVLC\_Refference\_Caffe (\textit{BVLC\_Ref}) model in Caffe \cite{Jia:2014:CCA:2647868.2654889} i.e. weight decay as 0.0005,  momentum as 0.9, and dropout of 0.5 is used in FC layers. In pre-processing nothing sophisticated has been done. 
We only used off-the-shelf normalization method available with Caffe. 
However, the learning parameters vary according to the dataset and respective state-of-the-art method being used in literature. 
 To calculate number of parameters for each layer we adopted 
number~of~parameters as $R1 \times R2 \times C \times M$, 
where $R1$ and $R2$ are rows and column size of kernel, $C$ is number of channels in that layer, and $M$ represents number of kernels for upper layer that will results in $M$ output feature maps.
\subsection{Impact of Pyramidal Structure}\label{sec:experAssertion}
Few standard \textit{CNN} architectures are examined for the impact of reversing and imposing a pyramidal structure to the networks. The optimal result for \textit{MNIST} and \textit{Cifar}-10 is shown respectively by \textit{Caffe\_LENET} and \textit{C10} in Table.~\ref{tab:refvsOurs}. 
\textit{Caffe\_LENET} has 20 filters in first convolutional layer (C1) and 50 filters in second convolutional layer (C2). If we reverse it i.e. $C1=50$ and $C2=20$, we call it \textit{SPyr\_Rev\_LENET} and it resulted in competitive accuracy as shown in Table.~\ref{tab:refvsOurs}. Same is the case for Cifar-10 i.e. \textit{SPyr\_Rev\_C10}. Cifar-100 consisted of 50 images per class consisting of 50000 training images. This makes it really challenging and we faced a 0.37\% loss in overall accuracy as shown in Table.~\ref{tab:refvsOurs}. \textit{SPyr\_Rev\_C100}. \textit{BVLC\_Ref} is similar to Alexnet but with a slight change in order of pooling and normalization layer \cite{Jia:2014:CCA:2647868.2654889}. \textit{BVLC\_Ref} model in reverse does not provide a pyramidal structure i.e. \textit{Rev\_BVLC\_Ref} has 256, 384, 384, 256, and 96 kernels at C1, C2, C3, C4, and C5, respectively. As expected, it does not learn. Therefore, we rearranged it, not only with the aim to provide a pyramidal structure but also maintaining the same number of filters. Hence, Table.~\ref{tab:refvsOurs}. \textit{SPyr\_Rev\_BVLC} gets 352 kernels in C1 and C2, followed by 256 kernels in C3 and C4. 
Finally, C5 layer gets 160 kernels due to adding 32 kernels from C1 and C2. As shown, reverse \textit{SPyr} models helped each standard \textit{CNN} model in increasing and retaining overall performance. 
However, in big networks we learned one important point:  
if the reverse of a model doesn't result in pyramidal structure, the performance will drop or the network will not learn at all; this is true in the case of Rev\_BVLC\_Ref model.\par
The question of how many kernels one should have at first layer or at each layer is still not fully theoretically asserted and is an open research problem. However, in our experiments with \textit{SPyr\_CNN's}, we experimented that the size of last 'conv' (C5) layer has high impact on overall accuracy.
Therefore, we concluded that there should be enough number of filters at first layer (C1), so that, after reduction in several layers, the last convolutional layer could get almost 40-60\% of the filters in C1. Otherwise, there would be a 1-3\% increase in error rate, or in worse case, it will not learn at all, e.g. \textit{Rev\_BVLC\_Ref}.
This is achieved reducing the number of kernels by a factor $'f' = 10, 15, 20\%$ from each layer to the next one, as shown in Table.~\ref{tab:redFactor}. This ensures pure pyramid structure unlike \textit{SPyr\_Rev\_BVLC} where we had two layers having the same number of kernels. We also tested the model on \textit{Cifar-10} and \textit{ImageNet}. \textit{Cifar-10} shows that with 10-15\% reductions in kernel number gets almost the same performance. Whereas, with a 20\% reduction, accuracy decreases by 0.86\%. In big models like ImageNet, our \textit{SPyr\_BVLC\_Ref**} with 10\% reduction in each layer including FC's layers improves accuracy by 0.59\% and 0.61\% compared to our \textit{SPyr\_BVLC\_Ref} and \textit{BVLC\_Ref}, respectively (seeTable.~\ref{tab:redFactor}). Whereas, models with 20\% reduction got bad results or did not learn due to few kernels in 'C5'.  
Further, \textit{BVLC\_Ref} \cite{Jia:2014:CCA:2647868.2654889} 
 results in maximum accuracy at 313000 iteration, whereas, we reported at 393000. However, at the end of 450000 iterations our \textit{SPyr\_BVLC\_Ref**} achieved 0.61\% accuracy. 
\begin{table}
\centering
MNIST \\
Caffe\_LENET = 20-50-500-10 \\
SPyr\_Rev\_LENET = 50-20-500-10 \\
\begin{tabular}{|l|c|c|c|}
\hline
Model & Train Loss & Test Loss & Accuracy \\
\hline\hline
Caffe\_LENET & 0.003735 & 0.029231 & 99.1 \\
SPyr\_Rev\_LENET &	0.003547 &	0.025142 &	99.13 \\
\hline
\end{tabular}\\
------ \\
CIFAR-10 \\
C10 = 32-32-64-10 \\
SPyr\_Rev\_C10 = 64-32-32-10 \\
\begin{tabular}{|l|c|c|c|}
\hline
Model & Train Loss & Test Loss & Accuracy \\
\hline\hline
C10 & 0.331029 &	0.532539 &	81.65 \\
SPyr\_Rev\_C10 &	0.321514 &	0.537438 &	81.67 \\
\hline
\end{tabular}\\
------ \\
CIFAR-100 \\
C100 = 150-170-200-100 \\
SPyr\_Rev\_C100 = 200-170-150-100 \\
\begin{tabular}{|l|c|c|c|}
\hline
Model & Train Loss & Test Loss & Accuracy \\
\hline\hline
C100 & 0.53676 &	1.5585 &	58.64 \\
SPyr\_Rev\_C100 &	0.63625 &	1.5549 &	58.27 \\
\hline
\end{tabular}\\
------ \\
ILSVRC-12 \\
BVLC\_Ref=96-256-384-384-256-4096-4096 \\
Rev\_BVLC\_Ref=256-384-384-256-96-4096-4096 \\
SPyr\_Rev\_BVLC=352-352-256-256-160-4096-4096 \\
\begin{tabular}{|l|c|c|c|}
\hline
Model & Train Loss & Test Loss & Accuracy \\
\hline\hline
BVLC\_Ref & 1.21952 &	1.840 &	57.03 \\
Rev\_BVLC\_Ref  &	Does &	Not &	Learn  \\
SPyr\_Rev\_BVLC  & 1.34997 &	1.842 &	56.81 \\
\hline
\end{tabular}\\
-----\\
\caption{Results for Referenced Models vs. their reversed model according to our statement}
\label{tab:refvsOurs}
\end{table}
\subsection{Parameter Reduction \& Size on Disk}\label{sec:experParams}
Another big impact of \textit{SPyr} approach is its reduction of parameters. It results in less size of trained model on disk as can be seen in Table.~\ref{tab:paramsAcc}. We did a number of experiments with \textit{MNIST} to see how much we can reduce parameters. At first glance, \textit{SPyr\_Rev\_LENET} reduced 55.5\% parameters and enhanced performance by 0.02\%. This reduction in parameters resulted in less space on disk i.e. with $431K$ parameters it needs $1.685 MB$ whereas, later it took 0.749MB space on disk. We examined it further and designed \textit{SPyr\_LENET} with $80\%$ reduction in parameters. It resulted in 0.03\% enhancement over \textit{Caffe\_LENET}. However, when we reduced more than 90\% of parameters from convolutional and FC layers, it retained same accuracy as shown in Table.~\ref{tab:paramsAcc}. \textit{SPyr\_LENET\textbf{*}}. 
\textit{SPyr\_LENET\textbf{**}} 
shows that if we concern about accuracy instead of memory space, 
than \textit{SPyr} models can perform much better. 
Similarly for Cifar-10 and Cifar-100, reverse models in Table.~\ref{tab:paramsAcc} not only reduces parameters but also gave competitive result. \par 
In case of ImageNet, the size of the \textit{BVLC\_Ref} trained model is $238MB$. Whereas, our \textit{SPyr\_BVLC\_Ref} and \textit{SPyr\_Rev\_BVLC} have $198MB$ for better results and $160MB$ for $1\%$ less accuracy, respectively. This is due to proper model selection having $10-20$ million fewer parameters resulting in reduction of ambiguity. Reduction is not only in 'Conv' layers but also from 'FC' due to less maps connected to neurons. 
These results showed that this approach can provide significant difference in specific real world application, where lack of memory storage is an issue. \par 
\begin{table}
\centering
MNIST \\
(A) Caffe\_LENET = 20-50-500-10 \\
(B) SPyr\_Rev\_LENET = 50-20-500-10 \\ 
(C) SPyr\_LENET = 35-15-500-10 \\
(D) SPyr\_LENET\textbf{*} = 25-15-100-10 \\
(E) SPyr\_LENET\textbf{**} = 100-68-100-10 \\
\begin{tabular}{|l|c|c|c|}
\hline
Model & Parameters & Size in MB & Accuracy \\
\hline\hline
(A) & 431080 & 	1.685MB &	99.1 \\
(B) &	191830 &	0.749MB &	99.13 \\
(C) & \textbf{48910} &	\textbf{0.191MB} &	\textbf{99.14} \\
(D) & \textbf{35150} &	\textbf{0.137MB} &	\textbf{99.1} \\
(E) & \textbf{219250} & \textbf{0.857MB} & \textbf{99.24} \\
\hline
\end{tabular}\\
------ \\
CIFAR-10 \\
(F) C10 = 32-32-64-10 \\
(G) SPyr\_Rev\_C10 = 64-32-32-10 \\
(H) SPyr\_C10 = 38-30-24-10 \\
(I) SPyr\_C10 = 84-64-44-10 \\
\begin{tabular}{|l|c|c|c|}
\hline
Model & Parameters & Size in MB & Accuracy \\
\hline\hline
(F) & 87978 &	0.335MB &	81.65 \\
(G) &	\textbf{83658} & \textbf{0.319MB} &	\textbf{81.67} \\
(H) &  51392 & 0.189MB & 80.9 \\
(I) & 214142 & 0.837MB & 83.04 \\
\hline
\end{tabular}\\
------ \\
CIFAR-100 \\
(J) C100 = 150-170-200-100 \\
(K) SPyr\_Rev\_C100 = 200-170-150-100 \\
(L) SPyr\_C100 = 200-128-64-100 \\
\begin{tabular}{|l|c|c|c|}
\hline
Model & Parameters & Size in MB & Accuracy \\
\hline\hline
(J) & 1811870 & 6.917 MB &	58.64 \\
(K) & 1733120 & 6.616MB &	58.27 \\
(L) & 952692 &	3.637MB &	57.23 \\
\hline
\end{tabular}\\ 
------ \\
ILSVRC-12 \\
(M) BVLC\_Ref \cite{Jia:2014:CCA:2647868.2654889}=96-256-384-384-256-4096-4096 \\
(N) SPyr\_BVLC\_Ref=384-320-256-224-192-4096-3840 \\
(O) SPyr\_Rev\_BVLC=352-352-256-256-160-4096-4096 \\
(P) SPyr\_BVLC\_Ref*=352-352-256-256-160-3840-2840 \\
(Q) SPyr\_BVLC\_Ref**=384-346-308-270-232-4096-3687-1000 \\
\begin{tabular}{|l|c|c|c|}
\hline
Model & Parameters & Size in MB & Accuracy \\
\hline\hline
(M) & 62378344 & 238.15MB & 57.03 \\
(N) & \textbf{54046920} & \textbf{198.26MB} & \textbf{57.05} \\
(O) & 43065448 & 160.93MB & 56.81  \\
(P) & 40867904 & 151.71MB & 56.40  \\
(Q) & 58741371 & 224.2MB  & \textbf{57.64} \\
\hline
\end{tabular}\\
-----\\
\caption{Parameters and their size on disk for base and pyramidal architectures along with their accuracies }
\label{tab:paramsAcc}
\end{table}
S. Han et al. \cite{learningConnections15} reduced parameters with a time consuming three step process. Rather than training weights, they learn those connections which are more important at first step, followed by pruning those connections, and retraining the remaining connection and their weights at final step. They achieved $99.23\%$ accuracy with $12x$ reduction in parameters but consumed $2x$ more time than normal, as shown in Table.~\ref{tab:paramsAcc}. Caffe\_LENET. In addition, they applied complex regularization and loss functions to find the optimal results. However, \textit{SPyr\_LENET\textbf{**}} achieved $99.24\%$ accuracy with about $50\%$ reduction in parameters and almost same amount of time (i.e. $47.6s\pm1$ on NVIDIA Quadro K4200 GPU) as \textit{Caffe\_LENET} (other simulations were also running on same computer). See for this Table.~\ref{tab:paramsAcc}. 
Further, our reduction is not only done in 'FC' layers but also in 'Conv' layers unlike the \textit{LeNet-5 Pruned} Table.~\ref{tab:paramsAcc}. (F) \cite{learningConnections15} which reduced parameters mainly in 'FC' layer. \textit{Alexnet\_Pruned} \cite{learningConnections15} achieved same result with $6.7M$ parameters, but their model took about $700K$ to $1000K$ iterations. Whereas, we achieved same results with simple techniques, $10M$ reduction in parameters and same number of $450K$ iterations. However, due to more maps in initial layer, our model takes slightly more time than \textit{BVLC\_Ref}.
In Fig.~\ref{fig:accuracyilsvrc1000}, D100\_Model\_1 shows one of our pyramidal models that reduced 30.9623\% parameters with only 0.0142\% degradation in performance compared to reference model. Table.~\ref{tab:BestSPyr} summarizes the best models interms of parameters and accuracy for \textit{MNIST, Cifar-10, Cifar-100} and \textit{ILSVRC-12}.
\begin{table}
\begin{center}
Cifar-10 \\
\begin{tabular}{|l|c|c|c|}
\hline
Model & $'f'$& Params & Accuracy \% \\
\hline\hline
(A) 64-58-52-10 & 10 & 245306 & 83.43 \\
(B) 64-54-44-10 & 15 & 197111 & 83.36 \\
(C) 64-52-40-10 & 20 & 158623 & 82.57 \\
\hline
\end{tabular}\\
------ \\
ILSVRC-12 \\
(D) SPyr\_BVLC\_Ref**= 384-346-308-270-232-4096-3687-1000 \\
(E) 384-346-308-270-232-4096-4096-1000 \\
(F) 384-326-270-212-156-4096-3687-1000 \\
(G) 384-326-270-212-156-4096-4096-1000 \\
(H) 384-308-232-156-80-4096-3277-1000 \\
\begin{tabular}{|l|c|c|c|c|}
\hline
Model&$'f'$& Params & top-1\% &top-5\% \\
\hline\hline
(D) & 10 & 58M & 57.64 & 80.68 \\
(E) & 10 & 60M & 57.62 & 80.64 \\
(F) & 15 & 46M & 56.38 & 79.67 \\
(G) & 15 & 48M & 56.61	& 79.76 \\
(H) & 20 & 33M & - & - \\
\hline
\end{tabular}
\end{center}
\caption{ Reduction in Kernels by Factor $'f'$ along with their accuracies}
\label{tab:redFactor}
\end{table}
\begin{table}
\centering
(A) MNIST\_Caffe\_LENET = 20-50-500-10 \\
(B) SPyr\_C10 = 52-42-32-10 \\ 
(C) SPyr\_Rev\_C100 = 200-170-150-100 \\
(D) SPyr\_BVLC\_Ref=384-320-256-224-192-4096-3840 \\
\begin{tabular}{|l|c|c|c|}
\hline
Model & Parameters & Train Loss & Accuracy \\
\hline\hline
(A) & 48910 &	0.00994945 & 99.14 \\
(B) & 94756	& 0.319729 & 81.94 \\ 
(C) & 1733120 & 0.445774 & 58.27 \\
(D) & 54046920 & 1.24036 & 57.05 \\
\hline
\end{tabular}\\
------ \\
\caption{ Best Strictly Pyramidal Models and their Accuracies}
\label{tab:BestSPyr}
\end{table}
\subsection{Performance of Pyramidal Models with Less data}
One of the questions raised against \textit{CNN} is that it doesn't work if small datasets are available. Therefore, we tested \textit{SPyr\_CNN's} with reduced data. We divided Cifar-100 and ILSVRC-12 in two new random training sets, i.e. $90\%$ and $80\%$ of the original one. 
Despite reduction of $150K$ and $300K$ randomly selected images for test1 and test2, respectively, we got only $1\%$ gradual degradation in overall accuracy (see Table.~\ref{tab:generalizationPower}). So, if we properly understand \textit{CNNs}, we can model such architectures in order to provide optimal results even with small training datasets. Our top-1 accuracy even with $90\%$ data is better than reported in \cite{MemBoundCNNCollinsK14} by $0.4\%$. However, when we reduced data other than $10\%$, the performance drops down below $0.7\%$. \par
\begin{figure}[!t]
\centering
\includegraphics[width=3in]{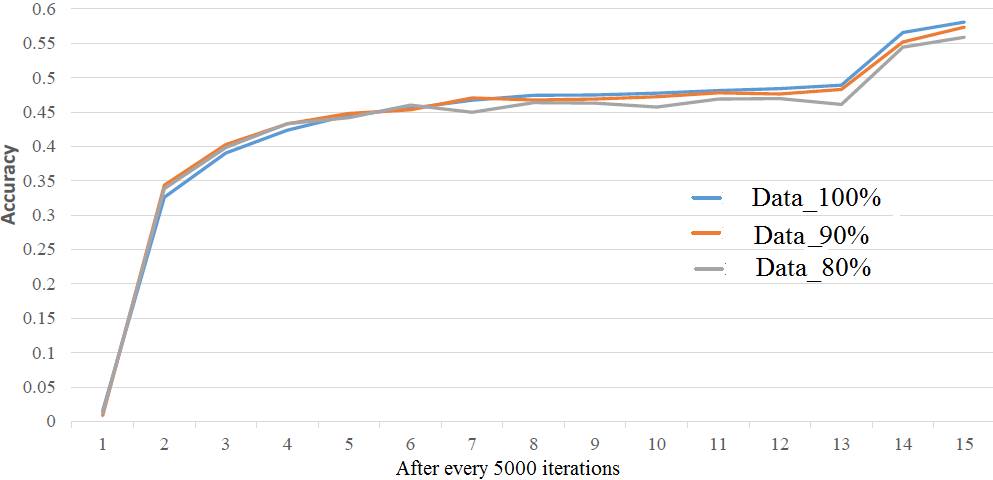}
   \caption{Performance evaluation based on Accuracy for CIFAR-100 with reduction in training data for total of 70000 iterations}
\label{fig:accuracyC100}
\end{figure}
\begin{figure}[!t]
\centering
\includegraphics[width=3in]{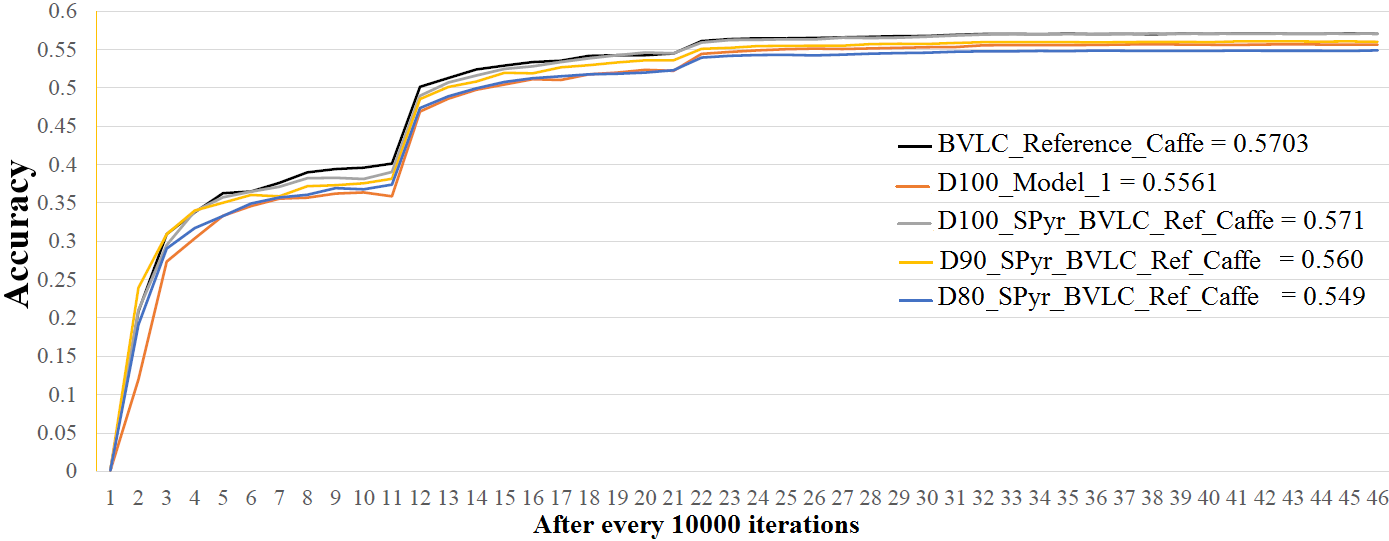}
   \caption{Comparison of Validation Accuracy with less data after each 10000 iterations for total of 450000 iterations}
\label{fig:accuracyilsvrc1000}
\end{figure}
\begin{figure}[!t]
\centering
\includegraphics[width=3in]{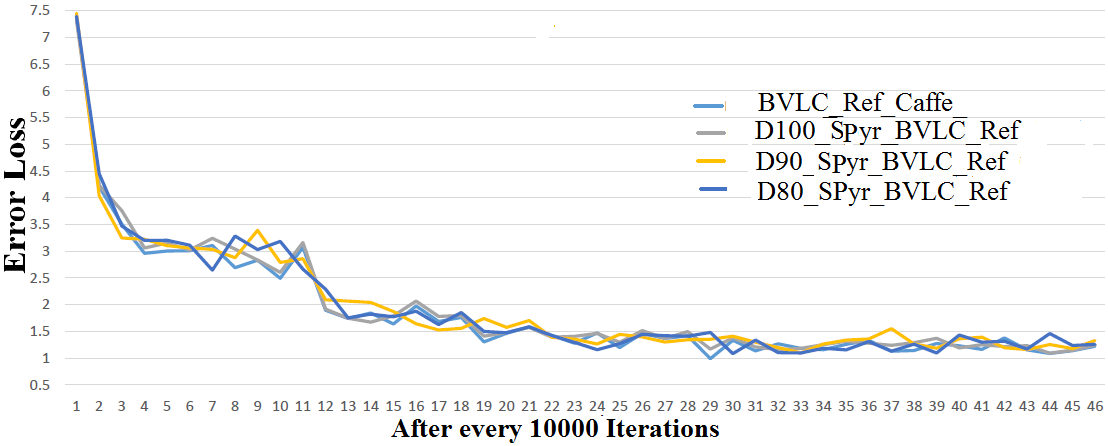}
   \caption{Training Loss with 10000 iterations difference for total of 450000 iterations}
\label{fig:trainloss}
\end{figure}

Performance of our \textit{SPyr\_CNN\_C100} with reduced data while training is shown in Fig.~\ref{fig:accuracyC100} (with 100\%, 90\% and 80\% training data). Not only the model with 100\% training data, but also the models with 90\% and 80\% data learn smoothly. Similarly, to see the training behavior of ILSVRC-12 with reduced data, Fig.~\ref{fig:trainloss} shows its learning behavior during training. Whereas, their performance through out the training process is shown in Fig.~\ref{fig:accuracyilsvrc1000}. The behavior for \textit{SPyr\_BVLC\_Ref} with 100\% (D100\_percent), 90\% (D90\_percent) and 80\% (D80\_percent) is almost smooth without any big falls or degradation. Curves shows almost similar nature despite less training data.  

\begin{table}
\centering
SPyr\_C100 = 200-170-128-100 \\
\begin{tabular}{|l|c|c|c|c|}
\hline
Data & Train Loss & Test Loss & top-1 & top-5\\
\hline\hline
100\% & 0.636249 & 1.57851 & 41.93 & - \\
90\% & 0.615743 & 1.64351	& 42.66 & - \\
80\% & 0.407365	& 1.72381 & 44.14 & - \\
\hline
\end{tabular}\\
------ \\
SPyr\_BVLC\_Ref=384-320-256-224-192-4096-3840 \\
\begin{tabular}{|l|c|c|c|c|}
\hline
Data & Train Loss & Test Loss & top-1 & top-5\\
\hline\hline
100\% & 0.449904 &	1.55493 & 42.9 &	19.7 \\
90\% & 1.32812 & 1.88455 & 44.0 &	20.7 \\
80\% & 1.25612 & 1.95334 & 45.1 & 21.4 \\
\hline
\end{tabular}\\
------ \\
\caption{ Evaluating generalization power of Strictly Pyramidal networks with reduction of training Data medium and large datasets. top-1 and top-5 represents error. }
\label{tab:generalizationPower}
\end{table}

\subsection{Comparison with State-of-the-art}
We compared the pyramidal structure mainly with Caffe reference models i.e. \textit{AlexNet} and \textit{BVLC\_Ref} as well as some state-of-the-art models as can be seen in Table.~\ref{tab:compStatofart}. 
Our \textit{SPyr\_LENET} model for MNIST outperformed based model as discussed in section \ref{sec:experParams}.
We achieved comparable results in case of Cifar-10 and Cifar-100 from the base models i.e. 81.65\% vs 83.34\% or 58.64\% vs 58.27\%. However, in comparison to models like \cite{DSNLeeXGZT15, NINLinCY13}, our results are $5-8\%$ less accurate. By observing their model, unlike ours, they contain a million parameters as well as results from sophisticated and complex models whereas, we achieved with following our assumption of reversing and imposing a \textit{SPyr} structure. Hence, if we neglect memory and increase parameters by increasing the \textit{SPyr} network, the performance improves. 
In case of large scale dataset, the results were quite promising in terms of accuracy. However, it took slightly more time due to more maps at initial layers. This limitation can be avoided by selecting a proper model. As \textit{SPyr\_LENET\textbf{*}} showed that \textit{SPyr} models can provide better or competitive results even with smaller number of kernels. In terms of reduction of parameters, recent models i.e. Table.~\ref{tab:compStatofart}. (F) \cite{MemBoundCNNCollinsK14} and (G) \cite{learningConnections15} reduced far more parameters than our models Table.~\ref{tab:compStatofart}. (C) and (D), 
 but they achieved less top-1 and top-5 error for ILSVRC-12.  \par

\begin{figure}[!t]
\centering
\includegraphics[width=3in]{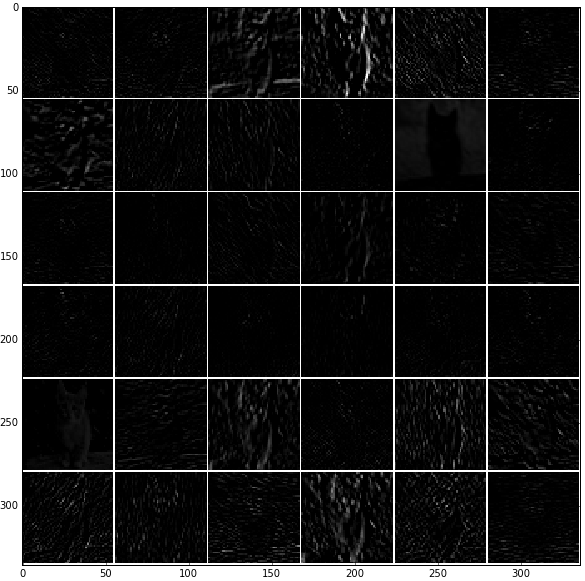}

   \caption{Output maps of first convolutional layer of Caffe trained model}
\label{fig:conv1data}
\end{figure}

\begin{figure}[!t]
\centering
\includegraphics[width=3in]{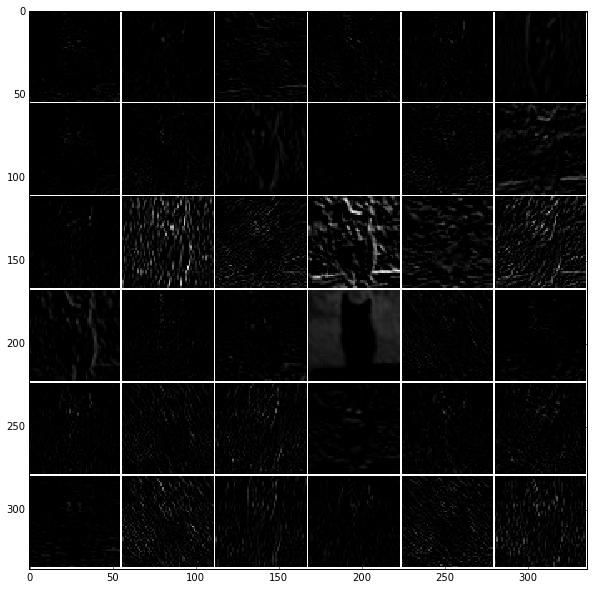}
   \caption{Output maps of first convolutional layer of our SPyr\_BVLC\_Ref trained model}
\label{fig:pyrconv1data}
\end{figure}

Some researchers are trying to avoid fully connected layers as majority of parameters are from those layers. Though, it should be noticed that these layers have great impact on over all accuracy. These are highly dependent on 2D or 3D layers. 
Therefore, one of the solution is to reduce the size of the last convolutional layer as-well-as number of neurons in specific order. Pyramidal structure is quite feasible for this scenario.  We used pyramid structure in \textit{FC} layers as can be seen in Table.~\ref{tab:paramsAcc}. (N), (O), and (Q). Table.~\ref{tab:paramsAcc}. (N) shows better result by following \textit{SPyr} structure while Table.~\ref{tab:paramsAcc}. (O) fallback with only $0.63\%$, whereas, Table.~\ref{tab:paramsAcc}. (Q) gives the best result by reducing kernels and neurons at a constant factor of 10.
To visualize and understand the output of our trained model, we have shown output maps resulted by first layer of trained \textit{BVLC\_Ref} model and our \textit{SPyr\_BVLC\_Ref} in Fig.~\ref{fig:conv1data} and \ref{fig:pyrconv1data}, respectively. The maps produced by our model are more clear, smooth and understandable as compare to the maps produced by referenced model. This shows that how better and fine grained information is extracted from real input image.
Still, a more detailed analysis is desired to precisely assess the effect of \textit{SPyr} architectures on new deep models and \textit{ImageNet} datasets. Such a comprehensive quantitative study using multiple networks is extremely time demanding and thus out of the scope of this paper.

\begin{table}
\begin{center}
\begin{tabular}{|l|c|c|}
\hline
Model Cifar-100 & Error \% \\
\hline\hline
C100 & 44.49 \\
Stochastic\_Pooling \cite{StochasticPoolZeiler} & 42.51 \\
SPyr\_C100 & 41.73 \\
NIN \cite{NINLinCY13} & 35.68 \\
Deeply Supervised Networks \cite{DSNLeeXGZT15} & 34.57 \\
\hline
\end{tabular}

------ \\
ILSVRC-12 \\
(A) Alexnet \cite{Jia:2014:CCA:2647868.2654889} \\
(B) BVLC\_Ref \cite{Jia:2014:CCA:2647868.2654889} \\
(C) SPyr\_BVLC\_Ref** \\
(D) SPyr\_BVLC\_Ref \\
(E) SPyr\_BVLC\_Ref* \\
(F) Sparse\_MemoryBounded \cite{MemBoundCNNCollinsK14} \\
(G) Alexnet\_Pruned \cite{learningConnections15} \\
\begin{tabular}{|l|c|c|c|c|c|}
\hline
Model & Params & top-1 \% & top-5\% \\
\hline\hline
(A) & 60M & 42.9 & 19.8 \\
(B) & 60M & 42.6 & 19.6 \\
(C) & 58M & 42.3 & \textbf{19.3} \\
(D) & 50M & 42.9 & 19.7 \\
(E) & 40M & 43.2 & 19.9 \\
(F) & 15M & 44.4 & 19.6 \\
(G) & 7M & 42.8 & 19.7 \\
\hline
\end{tabular}\\
\end{center}
\caption{Comparison with Stat-of-the-art in-terms of Error\% and Less number of Parameters}
\label{tab:compStatofart}
\end{table}

\section{Conclusion}\label{concl} 
We have demonstrated empirically that giving pyramidal structure to \textit{CNN's} can lead a scale down in the number of parameters as well as less solver memory consumption on disk, hence producing competitive results. 
Our experimental analysis was carried out on four standard datasets, which showed the effectiveness of the pyramidal structure. Training with reduced training data showed similar and smooth  learning with slight decrease in overall accuracy i.e. about $0.5-1\%$ decrease with each $10\%$ reduction. 
A suggestion for selecting number of kernels in each layer, especially first and last convolutional layer is given. 
In a sense, it makes it even more surprising that simple strict pyramidal model outperforms many existing sophisticated approaches. 

\bibliographystyle{IEEEtran}

\bibliography{egbib2}
\end{document}